\pdfoutput=1
\documentclass{article}
\usepackage{iclr2027_conference,times}

\usepackage{amsmath,amsfonts,bm}

\def\eqref#1{equation~\ref{#1}}

\def\1{\bm{1}}

\DeclareMathAlphabet{\mathsfit}{\encodingdefault}{\sfdefault}{m}{sl}
\SetMathAlphabet{\mathsfit}{bold}{\encodingdefault}{\sfdefault}{bx}{n}

\usepackage{float}
\usepackage{graphicx}
\usepackage{booktabs}
\usepackage{xcolor}
\usepackage{hyperref}
\usepackage{url}
\date{}
\iclrfinalcopy

\title{GroundingVLN: Reasoning and Acting with Grounding for Vision-Language Navigation}

\author{%
  \parbox[t]{\dimexpr\textwidth-2\tabcolsep\relax}{%
    \centering\normalfont
    \textbf{Kailing Li}\textsuperscript{1,*}\quad
    \textbf{Yu Han}\textsuperscript{1,3,*}\quad
    \textbf{Tianwen Qian}\textsuperscript{1, \dag}\quad
    \textbf{Yuqian Fu}\textsuperscript{2}\\[0.3ex]
    \textbf{Jingyu Gong}\textsuperscript{1}\quad
    \textbf{Jiangming Shi}\textsuperscript{1}\quad
    \textbf{Xiaoling Wang}\textsuperscript{1,\dag}\\[0.5ex]
    {\small
      \textsuperscript{1}School of Computer Science and Technology, East China Normal University\\
      \textsuperscript{2}King Abdullah University of Science and Technology\\[0.3ex]
      \textsuperscript{3}NeoteAI\\[0.3ex]
      \texttt{51275901046@stu.ecnu.edu.cn, twqian@cs.ecnu.edu.cn}\endgraf
    }%
  }%
}

\begin{document}

\maketitle
\lhead{Preprint}

\begingroup
\renewcommand{\thefootnote}{}
\footnotetext{\textsuperscript{*}Equal contribution.\quad
  \textsuperscript{\dag}Corresponding author.}
\endgroup

\begin{abstract}
Although vision-language models (VLMs) possess strong visual understanding and reasoning capabilities, vision-and-language navigation (VLN) agents built upon them still struggle to bridge semantic reasoning and spatial execution.
This disconnection stems from two coupled limitations: intermediate reasoning is not explicitly grounded in visual evidence, and high-level decisions lack precise spatial goals to guide low-level motion.
Inspired by human navigation, which naturally links cognition and locomotion through landmarks and spatial goals, we propose \textbf{GroundingVLN}, a framework that uses visual grounding as a shared interface between reasoning and action.
GroundingVLN first reasons with grounding by anchoring task-relevant visual evidence to precise image locations throughout structured reasoning.
It then acts through grounding by predicting a progress-aligned pixel goal, which is translated into primitive actions by a geometric planner.
To learn these capabilities, we construct \textbf{GroundingCOTVLN-188K}, a dataset of visually grounded reasoning traces synchronized with navigation progress, and introduce \textbf{Grounded and Execution-Aware Reinforcement Learning (GEAR)}, which aligns grounded reasoning and spatial decisions with downstream execution.
Experiments demonstrate that GroundingVLN achieves state-of-the-art performance (\textbf{69.9\%} SR on R2R-CE and \textbf{75.1\%} SR on RxR-CE) with superior data efficiency, requiring only \textbf{0.9\%} of the training data used by the strongest baseline.
It also generalizes strongly across datasets, attaining 59.9\% SR on RxR-CE when trained solely on R2R, a gain of 20.1\% over the strongest baseline.
Code is available at \url{https://github.com/Teacher-Tom/GroundingVLN}.
\end{abstract}

\section{Introduction}
\label{sec:introduction}

\begin{figure}[t]
    \centering
    \includegraphics[width=\linewidth]{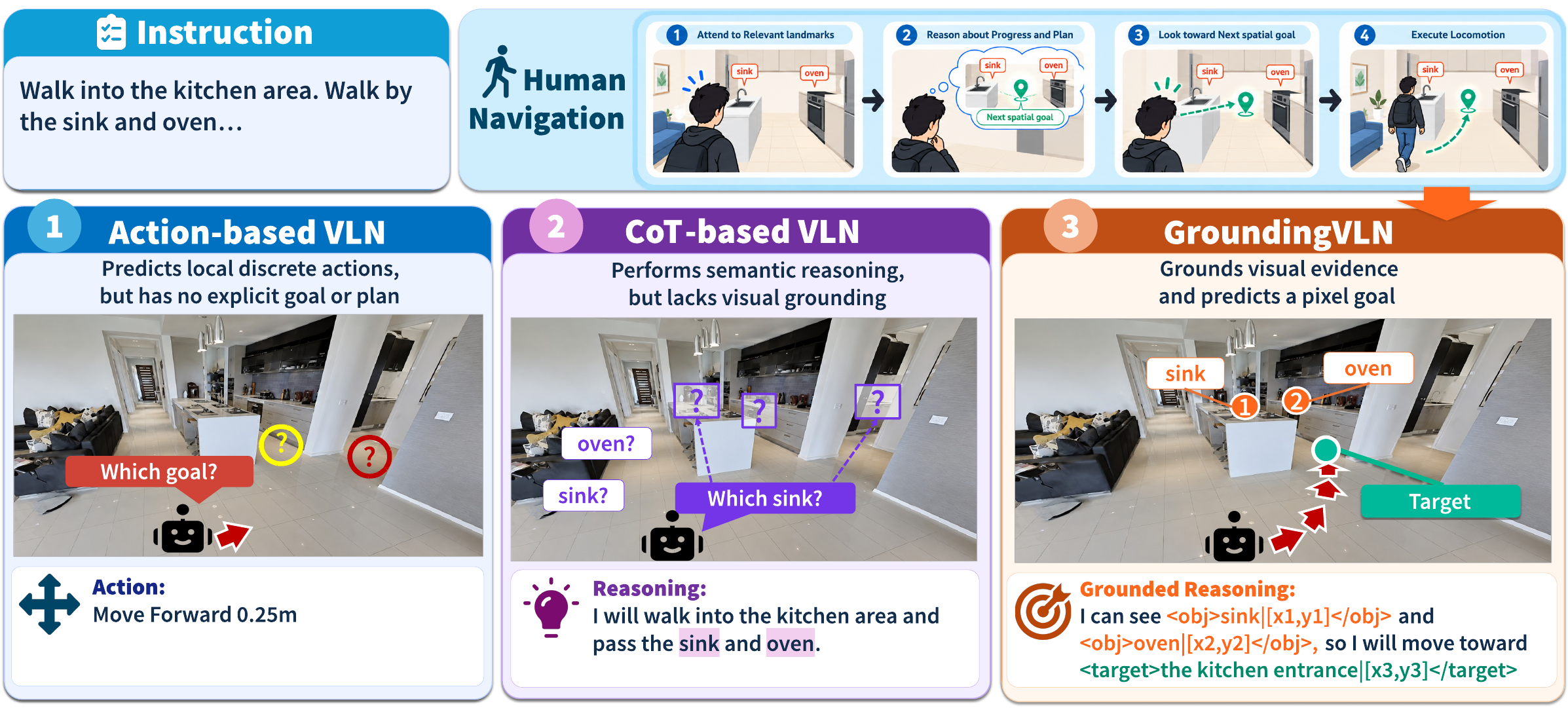}
    \caption{Human navigation grounds planning in relevant landmarks and spatial goals while separating cognition from motor control. Action-based VLN lacks explicit spatial goals, while text-only CoT leaves reasoning evidence ungrounded. GroundingVLN bridges these two coupled gaps by binding task-relevant evidence to image locations and predicting progress-aligned pixel goals, providing a shared grounding interface between high-level reasoning and low-level execution.}
    \label{fig:overview}
\end{figure}

Vision-and-Language Navigation (VLN) requires an embodied agent to follow natural-language instructions in continuous, partially observable environments~\citep{anderson2018vision,krantz2020beyond}. Successful navigation requires grounding language in egocentric observations, tracking instruction progress, planning sub-tasks, and executing actions. Although recent vision-language models (VLMs) exhibit strong visual understanding and reasoning capabilities~\citep{qwen2026qwen35,wang2025internvl35}, translating them into reliable long-horizon navigation remains challenging.

Existing methods still struggle to bridge high-level semantic reasoning and low-level action execution. A dominant line of work fine-tunes VLMs to directly predict discrete actions, such as moving forward, turning, and stopping~\citep{zhang2024navid,zhang2025uninavid,cheng2025navila}.
However, directly mapping visual observations to low-level actions offers limited interpretability and requires substantial navigation trajectories to learn motor control, rather than fully leveraging the pretrained reasoning capabilities of VLMs. Alternatively, waypoint-based methods select traversable candidates generated by an external waypoint predictor~\citep{shi2025smartway,qiao2025opennav}, avoiding direct low-level action prediction but restricting action space to predicted candidates, which may limit overall performance.
Recent approaches introduce Chain-of-Thought (CoT) reasoning to improve long-horizon decision-making and interpretability~\citep{guo2026awarevln,zuo2026fantasyvln}. Nevertheless, their intermediate reasoning relies primarily on text or latent features, making it difficult to bind referenced evidence to precise regions in the current view or to express an executable spatial target. Consequently, two coupled limitations persist: the reasoning process lacks explicit visual evidence  anchored through spatial grounding, and high-level decisions lack a geometrically precise, training-friendly interface to guide low-level action execution.

Cognitive science suggests that human navigation follows a hierarchical organization, in which higher-level cognitive systems plan routes and spatial goals~\citep{balaguer2016hierarchical,epstein2017cognitive}, while brainstem and spinal circuits execute locomotion~\citep{leiras2022brainstem}. Crucially, humans selectively encode relevant landmarks at decision points~\citep{janzen2004selective} and proactively direct their gaze toward upcoming landing targets before stepping~\citep{patla2003look}. Together, these findings motivate visual grounding as the interface between cognition and action: Grounding instruction-relevant landmarks can support localization, progress estimation, and planning, while grounding the next intended destination provides a spatial objective for low-level execution. Such an interface is feasible with modern VLMs, which already possess strong visual understanding, reasoning, and coordinate-level pointing capabilities~\citep{qwen2026qwen35,zhang2026thinking}. We therefore cast the VLM as a high-level cognitive system that grounds reasoning in visual evidence and decisions in spatial goals, leaving motor execution to a low-level planner.

Building on this insight, we propose \textbf{GroundingVLN}, a grounding-centric framework that integrates grounding during reasoning and decision making. During reasoning, the VLM follows a structured process of subtask decomposition, history recall, current perception, progress estimation, and future-goal planning, while binding key evidence in its reasoning chain to precise image locations. During decision making, it predicts a pixel goal aligned with the current semantic subtask, which a low-level planner converts into an executable trajectory and actions. To train these capabilities, we construct \textbf{GroundingCOTVLN-188K} for supervised finetuning and enforce temporal alignment of grounding across changing viewpoints. We further introduce \textbf{Grounded and Execution-Aware Reinforcement Learning (GEAR)}, which exploits the measurable geometry of pixel goals to provide fine-grained rewards that jointly align reasoning, spatial decisions, and downstream execution. Extensive experiments show that GroundingVLN achieves state-of-the-art performance on both R2R-CE and RxR-CE using only \textbf{0.9\%} of the training data used by the strongest baseline, while demonstrating strong cross-dataset generalization.

Our main contributions are as follows:
\begin{itemize}
    \item We introduce a grounding-centric VLN paradigm that decouples high-level cognition from low-level motion: grounding during reasoning binds intermediate thoughts to visual evidence, while grounding during decision making converts semantic subgoals into executable spatial goals.
    \item We build an automatic data engine to construct \textbf{GroundingCOTVLN-188K}, with a mechanism for temporal alignment of grounding that preserves grounded entities across changing viewpoints. We further propose \textbf{GEAR}, which exploits the continuous geometry of pixel goals to provide fine-grained rewards that align reasoning, spatial decisions, and execution.
    \item Extensive experiments demonstrate that GroundingVLN achieves state-of-the-art performance on R2R-CE and RxR-CE with only \textbf{0.9\%} of the training data used by the strongest baseline, while exhibiting strong data efficiency and cross-dataset generalization.
\end{itemize}

\section{Related Work}
\label{sec:related-work}

\subsection{Action Representations in Vision-and-Language Navigation}
\label{sec:related-action}

Recent VLN systems increasingly adapt pretrained VLMs as navigation policies. NaVid, Uni-NaVid, and StreamVLN directly generate low-level actions~\citep{zhang2024navid,zhang2025uninavid,wei2026streamvln}, while NaVILA outputs language-form mid-level actions for a locomotion policy~\citep{cheng2025navila}. Such action prediction offers a simple interface, but provides limited supervision for task progress and visual evidence, often requires substantial training data, and may compromise the VLM's general-purpose capabilities through catastrophic forgetting~\citep{hancock2026actions}. Waypoint-based methods instead select among traversable candidates proposed by a dedicated predictor~\citep{krantz2021waypoint,an2025etpnav,shi2025smartway}; although avoiding direct low-level action prediction, their action space and performance remain bounded by candidate quality. DualVLN further predicts the farthest visible pixel goal, but conditions its diffusion executor on latent VLM features~\citep{wei2026dualvln}. In contrast, GroundingVLN uses a progress-aligned pixel goal as an explicit interface between semantic reasoning and low-level planning.

\subsection{Chain-of-Thought Reasoning in Vision-and-Language Navigation}
\label{sec:related-vln-cot}

To improve interpretability and long-horizon decision making, NavGPT-2 and NavCoT generate textual chains of thought before predicting actions~\citep{zhou2024navgpt2,lin2025navcot}. However, free-form reasoning can be verbose, hallucinate visual details, and increase inference cost. Recent methods therefore adopt more structured reasoning: EvolveNav combines formalized CoT supervision with self-improvement~\citep{lin2026evolvenav}, AwareVLN reasons about agent state and task progress at key decision points~\citep{guo2026awarevln}, and FantasyVLN incorporates visual imagination into multimodal CoT~\citep{zuo2026fantasyvln}. Nevertheless, their intermediate conclusions are not explicitly bound to image regions, making spatial descriptions difficult to verify against current observations. GroundingVLN instead introduces coordinate-based grounding into structured CoT, directly linking each reasoning step to its visual evidence.

\subsection{Reasoning with Grounding}
\label{sec:related-grounded-reasoning}

Multimodal reasoning is shifting from text-only CoT toward explicit interaction with visual evidence. Visual CoT Prompting iterates through seeing, reasoning, and verification~\citep{chen2024visualcot}, while V* introduces guided visual search for task-relevant regions~\citep{wu2024vstar}. More recent methods directly ground intermediate thoughts: ViGoRL anchors each reasoning step to image coordinates~\citep{sarch2025grounded}, and Thinking with Visual Grounding interleaves language with point or box references~\citep{zhang2026thinking}. However, these methods primarily address single-image reasoning and cannot preserve entity identity across changing viewpoints. GroundingVLN extends grounded reasoning to closed-loop navigation through temporally aligned grounding, allowing visual evidence to be tracked and reused throughout a trajectory.

\section{Method}
\label{sec:method}

\subsection{Problem Formulation}
\label{sec:problem-formulation}

We consider vision-and-language navigation in an unknown continuous 3D environment. Given a natural-language instruction $I=(w_1,\ldots,w_L)$, an agent starts from state $s_0$ and receives an egocentric visual observation $o_t$ at each time step $t$. Based on the instruction, current observation, and interaction history $\mathcal{H}_t=\{(o_\tau,a_\tau)\}_{\tau=0}^{t-1}$, the policy $\pi_\theta$ produces a navigation decision $a_t \sim \pi_\theta(a_t \mid I,o_t,\mathcal{H}_t)$.
The agent must reach the destination and stop within a finite horizon.

\begin{figure}[t]
    \centering
    \includegraphics[width=\linewidth]{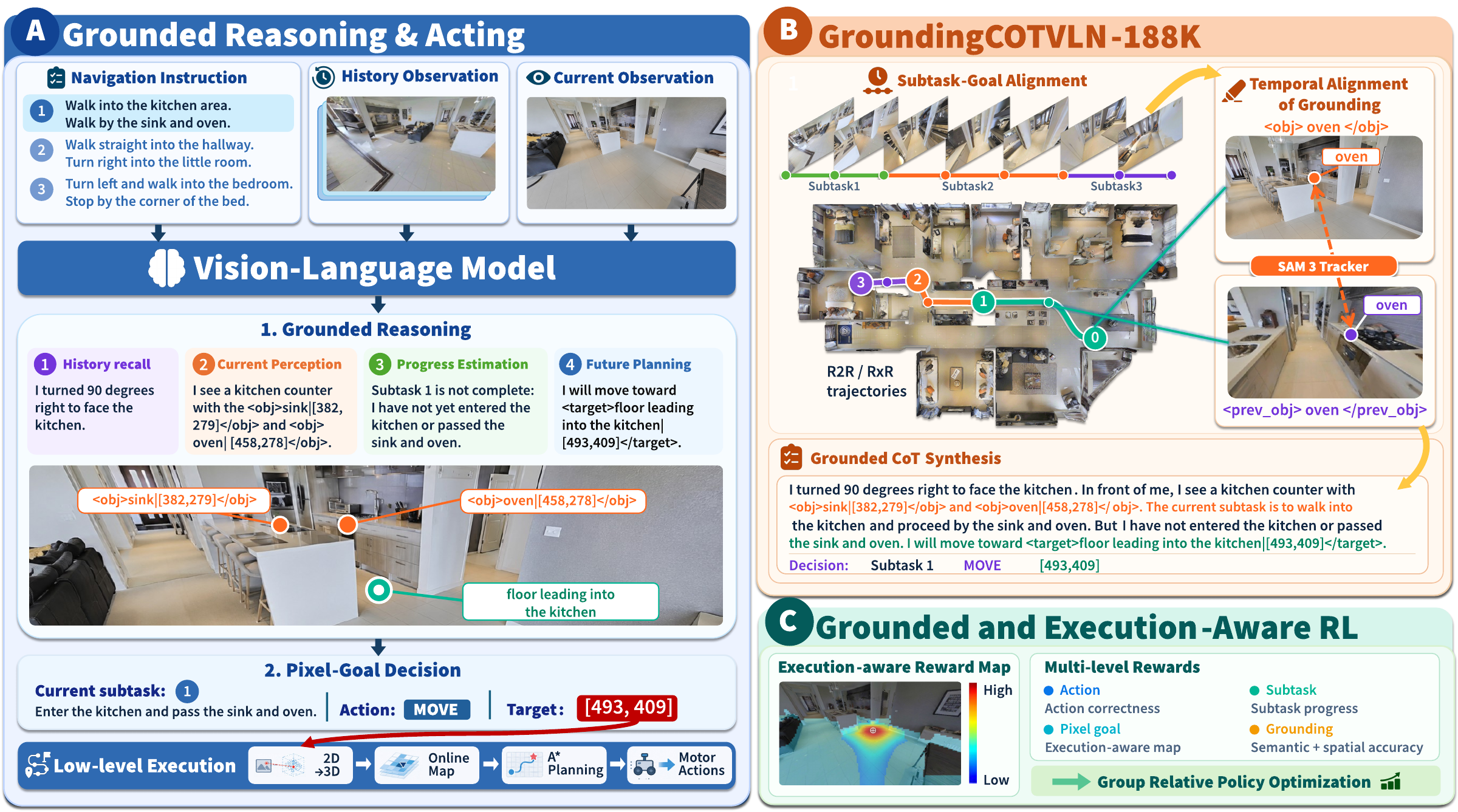}
    \caption{\textbf{Overview of GroundingVLN.} (A) The instruction is first decomposed into ordered subtasks. The high-level VLM performs grounded reasoning and predicts the current subtask, action, and target, which the low-level execution module converts into primitive actions. (B) GroundingCOTVLN-188K provides grounded CoT supervision through subtask--goal alignment and temporal alignment of grounding. (C) GEAR combines an execution-aware reward map with multi-level rewards for group-relative policy optimization.}
    \label{fig:framework}
\end{figure}

\subsection{Framework of GroundingVLN}
\label{sec:framework}

Figure~\ref{fig:framework} summarizes GroundingVLN's decoupled, closed-loop design. (A) After decomposing the instruction into ordered subtasks, the VLM performs structured, visually grounded reasoning to recall past execution, perceive the scene, assess progress, and plan the next goal. It anchors relevant evidence with coordinate-tagged \texttt{<obj>} and \texttt{<prev\_obj>} references. It then predicts the current subtask, a high-level action (\texttt{MOVE}, \texttt{TURN}, or \texttt{STOP}), and a target pixel as a decision. For \texttt{MOVE}, the low-level module back-projects the 2D pixel goal into 3D, plans a path, and executes it. (B) To train grounded reasoning and pixel-goal decision-making, we construct the GroundingCOTVLN-188K dataset. Subtask--goal alignment ensures that each decision's pixel goal matches subtask progress, while temporal alignment of grounding keeps visual references consistent across observations during reasoning. (C) GEAR (Grounded and execution-aware reinforcement learning) combines an execution-aware pixel-goal reward map with multi-level rewards for group-relative policy optimization, further improving grounding and decision-making.

\subsection{Reasoning and Acting with Grounding}
\label{sec:reasoning-acting}

At each decision step, GroundingVLN first binds evidence to image locations during structured reasoning, then grounds the resulting decision to a pixel goal. A low-level planner converts this goal into primitive actions, closing the loop between reasoning, decision making, and execution.

\paragraph{Structured Reasoning with Grounding.}
\label{sec:structural-reasoning}

Inspired by the hierarchical and goal-directed nature of human navigation, we decompose the instruction once at initialization into ordered subtasks, identifying the turns and visual evidence required by each. At each subsequent decision step, the VLM performs four ordered reasoning stages:
\begin{enumerate}
    \item \textbf{History Recall:} Summarize the previous execution, including traveled distance, heading changes, and observed landmarks.
    \item \textbf{Current Perception:} Describe the current scene and orientation while identifying visible instruction-relevant evidence.
    \item \textbf{Progress Estimation:} Determine which requirements have been satisfied, whether the current subtask is complete, and what remains unresolved.
    \item \textbf{Future Planning:} Select the next local objective according to the earliest unfinished instruction condition.
\end{enumerate}

To bind this reasoning process to the observation, we embed visual grounding directly into the CoT. A newly confirmed entity is represented as \texttt{<obj>l|[x,y]</obj>}, whereas a previously grounded entity that reappears in the current view is marked as \texttt{<prev\_obj>l|[x,y]</prev\_obj>}. Here, $l$ denotes a semantic label, and all grounding and target locations use normalized integer coordinates $(x,y)\in[0,1000]^2$ in the current view. These references make each visual claim spatially verifiable and allow historical evidence to be explicitly reused during progress estimation and planning.

\paragraph{Progress-Aligned Pixel-Goal Decision.}
\label{sec:pixel-goal-decision}

After structured reasoning, the VLM predicts a high-level action $a_t^h\in\{\texttt{MOVE},\texttt{TURN},\texttt{STOP}\}$. A \texttt{MOVE} always includes a pixel goal $g_t=(c_t,l_t,p_t)$, specifying the current subtask $c_t$, target label $l_t$, and location $p_t=(x_t,y_t)$. Its walkable destination is marked in the CoT as \texttt{<target>l|[x,y]</target>}, distinct from the evidential landmarks denoted by \texttt{<obj>} and \texttt{<prev\_obj>}. A \texttt{TURN} specifies $d_t\in\{\texttt{left},\texttt{right},\texttt{around}\}$ and includes the same target tag and pixel goal only when its intended post-turn destination is visible in the current observation. The current step executes only an in-place rotation of $90^\circ$ for \texttt{left}/\texttt{right} or $180^\circ$ for \texttt{around}. \texttt{STOP} is emitted only when all instruction conditions are satisfied.

Unlike DualVLN, which supervises the farthest visible trajectory point~\citep{wei2026dualvln}, we align the pixel goal with semantic progress. We partition the reference trajectory into ordered subtask segments and restrict candidates at step $t$ to the current segment, terminating at its next decision boundary. Among candidates that are continuously visible and traversable, we select the point with the greatest forward path progress and project it into the current view. The model jointly predicts $(c_t,l_t,p_t)$ at inference, coupling what remains to be completed with where the agent should move next without prematurely crossing into a future subtask.

\paragraph{Low-Level Planning and Execution.}
\label{sec:low-level-execution}

The execution module handles each decision according to its action type: \texttt{TURN} triggers a fixed in-place rotation, with or without a pixel goal, and \texttt{STOP} ends navigation. Only \texttt{MOVE} uses its pixel goal to plan and execute a path. RTAB-Map~\citep{labbe2019rtabmap} estimates the camera pose online from the RGB-D stream, while accumulated point clouds form a map of ground support, obstacles, and elevation changes, including stairs. For each \texttt{MOVE}, the module converts the target to local pixel coordinates, then uses the corresponding depth, camera intrinsics, and estimated pose to recover its 3D world position. A* searches the map for a collision-free path to this position, and a path follower translates the path into low-level turns and forward movements.

\subsection{GroundingCOTVLN-188K: Automatic Grounded SFT Data Generation}
\label{sec:sft-data-engine}

To train structured reasoning and pixel-goal prediction, we construct \textbf{GroundingCOTVLN-188K}, comprising 188K grounded image--CoT samples collected from 21K R2R and RxR trajectories.

\paragraph{Trajectory Collection.}
\label{sec:trajectory-collection}

For each instruction--reference-path pair, we replay the trajectory in Habitat~\citep{savva2019habitat}. Densely spaced reference points are merged, long segments are interpolated, and the low-level planner drives the agent between successive points while recording its actual executed trajectory. We sample three types of decision anchors: \emph{arrival anchors} at reference points, \emph{turn anchors} around substantial in-place rotations, and \emph{mid-distance anchors} after sufficient travel. Additional anchors are inserted across large gaps, whereas near-static or redundant samples are removed. Each retained anchor stores stitched left-front-right RGB-D observations, camera pose, reference-point index, and executed position, providing the geometric basis for progress annotation, visual projection, and pixel-goal construction.

\paragraph{Subtask--Goal Alignment.}
\label{sec:subtask-target-alignment}

We first use the teacher VLM to decompose the instruction into ordered subtasks $\{u_k\}_{k=1}^{K}$ and extract landmarks and turns. Using replayed observations and reference actions, each $u_k$ is aligned with a contiguous path segment $\mathcal{S}_k$, allowing every anchor $t$ to be assigned a current progress segment $\mathcal{S}_{k_t}$ and a set of completed subtasks.

Pixel-goal candidates are further constrained by actual execution between the current and next anchors. Let $\mathcal{T}_{t:t+1}$ denote this executed trajectory segment; we select
\begin{equation}
q_t^*=\underset{q\in\mathcal{T}_{t:t+1}\cap\mathcal{S}_{k_t}}{\arg\max}\;\operatorname{Prog}(q),
\quad \mathrm{s.t.}\quad
\operatorname{Visible}(q,o_t)\wedge\operatorname{Navigable}(q),
\end{equation}
where $\operatorname{Prog}(q)$ measures forward progress along the path, and $\operatorname{Visible}(q,o_t)$ denotes continuous visibility along the trajectory up to $q$ in $o_t$. Depth-based occlusion checks ensure that $q_t^*$ lies on a genuinely visible ground region, after which camera projection produces $p_t=\Pi_t(q_t^*)\in[0,1000]^2$. This construction prevents targets from crossing into a future subtask while preserving their physical executability.

\paragraph{Grounded CoT Synthesis.}
\label{sec:grounded-cot-synthesis}

For each anchor, annotation proceeds in two stages. The teacher VLM first describes the agent orientation, spatial layout, visible landmarks, and task-relevant evidence in the current observation; it then performs directed grounding over referenced entities and inserts \texttt{<obj>l|[x,y]</obj>} tags.

Unlike grounding in a static image, VLN requires temporal alignment of grounding as the viewpoint changes. The same entity moves across image coordinates and may become occluded, creating two risks: grounding may drift to a similar instance, or a temporarily invisible entity may be forgotten. We therefore back-project every new grounding into a uniquely identified 3D world marker and initialize SAM 3 tracking~\citep{carion2026sam3}. At later anchors, historical markers are first reprojected into the current view; when reprojection is invalid, SAM 3 recovers the instance mask and valid depth points are back-projected to update its world position. A track is accepted only when its new 3D estimate satisfies spatial continuity with the stored marker. Invisible entities retain their identity in memory and, when observed again, are rebound to the current pixel location using \texttt{<prev\_obj>}.

Finally, deterministic programs assemble executed motion, current observation, progress state, action, and pixel goal, while the teacher VLM verbalizes these facts as a structured CoT. Samples are retained only if they pass coordinate visibility, depth occlusion, action--reasoning consistency, format completeness, and annotation-leakage checks, yielding the final GroundingCOTVLN-188K dataset. Appendix~\ref{sec:appendix-data-engine} provides further implementation details.

\subsection{Grounded and execution-aware reinforcement learning}
\label{sec:gear}

The continuous and measurable geometry of pixel goals enables finer post-training signals than discrete action correctness alone. We therefore introduce \textbf{GEAR} (Grounded and execution-aware reinforcement learning), which constructs execution-aware 2D goal reward maps and combines them with multi-level rewards through Group Relative Policy Optimization (GRPO)~\citep{shao2024deepseekmath}. Appendix~\ref{sec:appendix-reward-definitions} gives the exact reward definitions, and Appendix~\ref{sec:appendix-grpo} provides the optimization details.

\paragraph{Execution-Aware Pixel-Goal Reward Maps.}
\label{sec:goal-reward-map}

Rewarding only the 2D distance to a reference pixel ignores its mismatch with physical distance and cannot reflect obstacles or terrain. For each anchor $t$, we therefore construct a reward map $\mathcal{M}_t$ over the image domain. A candidate pixel $p$ is back-projected to obtain an execution endpoint $P_t(p)$; endpoints on walls, furniture, or other non-traversable surfaces are removed using the reconstructed local traversability map. For each valid candidate, we define
\begin{equation}
Q_t(p)=
\lambda_r\exp\!\left(-\frac{d_{\perp}(p)^2}{2\sigma_r^2}\right)
+\lambda_p\exp\!\left(-\frac{(s(p)-s_t^*)^2}{2\sigma_p^2}\right)
+\lambda_e\exp\!\left(-\frac{d_e(p)^2}{2\sigma_e^2}\right),
\end{equation}
where $d_{\perp}(p)$ is the lateral distance to the reference route, $s(p)-s_t^*$ is the route-progress difference from the reference target, and $d_e(p)=\lVert P_t(p)-P_t^*\rVert_2$ measures execution-endpoint error. Thus, $Q_t$ jointly scores route consistency, semantic progress, and physical executability; invalid pixels receive the minimum map value. Figure~\ref{fig:framework}(C) visualizes the resulting spatial distribution of $Q_t$.

\paragraph{Multi-Level Navigation Reward and Group-Relative Policy Optimization.}
\label{sec:multi-level-grpo}

Beyond the pixel-goal reward map, we design complementary rewards for output validity, action selection, subtask progress, and visual grounding. Invalid outputs receive the minimum reward; otherwise, for each action--target case $y$, we compute
\begin{equation}
R_y=\alpha_y^{a}R_{\mathrm{act}}
+\alpha_y^{t}R_{\mathrm{task}}
+\alpha_y^{g}R_{\mathrm{goal}}
+\lambda_gR_{\mathrm{grd}}-P,
\qquad
R_{\mathrm{goal}}(p)=\phi\!\left(Q_t(p)\right),
\end{equation}
where $R_{\mathrm{act}}$, $R_{\mathrm{task}}$, and $R_{\mathrm{grd}}$ evaluate the action, current subtask, and the semantic and spatial correctness of grounded evidence, respectively; $P$ penalizes inconsistencies between reasoning and decisions. For each anchor, we sample $G$ outputs and optimize their group-normalized rewards with a clipped group-relative policy objective, encouraging responses that jointly improve reasoning, grounding, and executable navigation.

\begingroup
\let\arxivtable\table
\def\arxivEarlyTables{}
\let\arxivfigure\figure
\renewcommand{\topfraction}{0.7}
\renewcommand{\textfraction}{0.25}
\renewcommand{\table}[1][]{%
  \ifnum\value{table}=0
    \arxivtable[!b]
  \else
    \ifnum\value{table}=2
      \arxivtable[t]
    \else
      \arxivtable[!t]
    \fi
  \fi
}
\renewcommand{\figure}[1][]{%
  \ifnum\value{figure}=3
    \arxivfigure[!t]
  \else
    \arxivfigure[t]
  \fi
}
\ifdefined\arxivEarlyTables
\begin{table}[htbp]
\centering
\caption{Comparison on R2R-CE and RxR-CE \texttt{val\_unseen}. Data reports the number of individual training samples stated by each method; counting details are provided in Appendix~\ref{sec:appendix-data-efficiency}. Params reports backbone scale, excluding additional learned components (+). $\dagger$ includes 13.1M samples inherited from NaVILA and 1.9M AwareVLN-specific samples. $\ddagger$ comprises 16.9M expert trajectory samples and 5.0M reasoning samples.}
\label{tab:main-results}
\begingroup
\footnotesize
\setlength{\tabcolsep}{1.6pt}
\renewcommand{\arraystretch}{1.05}
\resizebox{\textwidth}{!}{%
\begin{tabular}{@{}lcccccccccc@{}}
\toprule
\raisebox{-0.6\baselineskip}{Method} & \raisebox{-0.6\baselineskip}{Data} & \raisebox{-0.6\baselineskip}{Params} & \multicolumn{4}{c}{R2R-CE Val-Unseen} & \multicolumn{4}{c}{RxR-CE Val-Unseen} \\
\cmidrule(lr){4-7}\cmidrule(lr){8-11}
 & & & NE$\downarrow$ & OS$\uparrow$ & SR$\uparrow$ & SPL$\uparrow$ & NE$\downarrow$ & SR$\uparrow$ & SPL$\uparrow$ & nDTW$\uparrow$ \\
\midrule
HPN+DN~\citet{krantz2021waypoint} & -- & -- & 6.31 & 40.0 & 36.0 & 34.0 & -- & -- & -- & -- \\
CMA~\citet{hong2022bridging} & -- & -- & 6.20 & 52.0 & 41.0 & 36.0 & 8.76 & 26.5 & 22.1 & 47.0 \\
GridMM~\citet{wang2023gridmm} & -- & -- & 5.11 & 61.0 & 49.0 & 41.0 & -- & -- & -- & -- \\
ETPNav~\citet{an2025etpnav} & -- & -- & 4.71 & 65.0 & 57.0 & 49.0 & 5.64 & 54.7 & 44.8 & 61.9 \\
ScaleVLN~\citet{wang2023scalevln} & 4.9M & -- & 4.80 & -- & 55.0 & 51.0 & -- & -- & -- & -- \\
\midrule
InstructNav~\citet{long2024instructnav} & 0 & -- & 6.89 & -- & 31.0 & 24.0 & -- & -- & -- & -- \\
SmartWay~\citet{shi2025smartway} & 0 & -- & 7.01 & 51.0 & 29.0 & 22.5 & -- & -- & -- & -- \\
OpenNav~\citet{qiao2025opennav} & 0 & -- & 6.70 & 23.0 & 19.0 & 16.1 & -- & -- & -- & -- \\
HSGM~\citet{li2026hsgm} & 0 & -- & 5.42 & 58.7 & 47.9 & 32.8 & 7.43 & 41.8 & 25.1 & 54.9 \\
\midrule
VLN-R1~\citet{qi2025vlnr1} & 1.83M & 7B & 7.00 & 41.2 & 30.2 & 21.8 & 9.10 & 22.7 & 17.6 & -- \\
NaVid~\citet{zhang2024navid} & 1.28M & 7B & 5.47 & 49.1 & 37.4 & 35.9 & -- & -- & -- & -- \\
NaVILA~\citet{cheng2025navila} & 13.1M & 8B & 5.22 & 62.5 & 54.0 & 49.0 & 6.77 & 49.3 & 44.0 & 58.8 \\
Uni-NaVid~\citet{zhang2025uninavid} & 3.58M & 7B & 5.58 & 53.5 & 47.0 & 42.7 & 6.24 & 48.7 & 40.9 & -- \\
StreamVLN~\citet{wei2026streamvln} & $\sim$26.3M & 7B & 4.98 & 64.2 & 56.9 & 51.9 & 6.22 & 52.9 & 46.0 & 61.9 \\
NavFoM~\citet{zhang2025navfom} & 12.7M & 7B & 4.61 & 72.1 & 61.7 & 55.3 & 4.74 & 64.4 & 56.2 & 65.8 \\
MapDream~\citet{lian2026mapdream} & 1.7M & 2B+ & 4.59 & 64.4 & 59.8 & 54.4 & 4.96 & 59.4 & 49.2 & -- \\
DualVLN (S2)+SPF~\citet{wei2026dualvln} & $\sim$12.0M & 7B & 4.25 & 68.3 & 60.9 & 55.2 & 5.71 & 63.5 & 55.0 & 46.8 \\
DualVLN~\citet{wei2026dualvln} & $\sim$12.0M & 7B+ & 4.05 & 70.7 & 64.3 & 58.5 & 4.58 & 61.4 & 51.8 & 70.0 \\
JanusVLN~\citet{zeng2026janusvln} & 10.69M & 7B & 4.78 & 65.2 & 60.5 & 56.8 & 6.06 & 56.2 & 47.5 & 62.1 \\
AwareVLN~\citet{guo2026awarevln} & 15.0M$^{\dagger}$ & 8B & 4.02 & 73.5 & 65.4 & 55.1 & 3.95 & 67.6 & 56.1 & 65.7 \\
ABot-N0~\citet{chu2026abotn0} & 21.9M$^{\ddagger}$ & 4B+ & 3.78 & 70.8 & 66.4 & 63.9 & 3.83 & 69.3 & 60.0 & -- \\
\midrule
\textbf{GroundingVLN (Ours)} & \textbf{188K} & \textbf{4B} & \textbf{3.66} & \textbf{74.8} & \textbf{69.9} & \textbf{64.1} & \textbf{3.54} & \textbf{75.1} & \textbf{62.0} & \textbf{75.3} \\
\bottomrule
\end{tabular}
}
\endgroup
\end{table}

\fi

\section{Experiments}
\label{sec:experiments}

\subsection{Experimental Setup}
\label{sec:experimental-setup}

\paragraph{Benchmarks.}
We evaluate GroundingVLN on R2R-CE and RxR-CE under the continuous-environment VLN setting~\citep{anderson2018vision,krantz2020beyond,ku2020rxr}. Both benchmarks are implemented in Habitat~\citep{savva2019habitat} using scenes from Matterport3D~\citep{chang2017matterport3d}, and we report results on their \texttt{val\_unseen} splits. Following standard protocols~\citep{anderson2018evaluation,ilharco2019ndtw}, we report Navigation Error (NE), Oracle Success Rate (OS), Success Rate (SR), and Success weighted by Path Length (SPL), where success requires stopping within $3\,\mathrm{m}$ of the goal; for RxR-CE, we additionally report normalized Dynamic Time Warping (nDTW).

\paragraph{Implementation Details.}
We use Qwen3.5-397B-A17B~\citep{qwen2026qwen35} to annotate GroundingCOTVLN-188K and initialize GroundingVLN from Qwen3.5-4B. Training consists of two stages. We first perform one epoch of supervised fine-tuning on 188K grounded image--CoT samples covering 21K R2R and RxR trajectories, which takes approximately 8 hours on 8 NVIDIA H200 GPUs. We then apply GEAR to 16K samples for 2,000 optimization steps with a learning rate of $1\times10^{-6}$, taking approximately 20 hours on the same hardware.

\ifdefined\arxivEarlyTables\else

\fi

\ifdefined\arxivEarlyTables
\begin{table}[htbp]
\centering
\caption{R2R-to-RxR transfer on the RxR-CE \texttt{val\_unseen} split without using RxR-CE training data.}
\label{tab:cross-dataset-generalization}
\begingroup
\small
\setlength{\tabcolsep}{7pt}
\renewcommand{\arraystretch}{1.05}
\begin{tabular}{@{}lcccc@{}}
\toprule
Method & NE$\downarrow$ & OS$\uparrow$ & SR$\uparrow$ & SPL$\uparrow$ \\
\midrule
NaVid~\citet{zhang2024navid} & 8.41 & 34.5 & 23.8 & 21.2 \\
MonoDream~\citet{wang2026monodream} & 8.57 & 35.9 & 25.1 & 21.6 \\
Progress-Think~\citet{wang2025progressthink} & 8.30 & 38.3 & 27.5 & 22.7 \\
NaVILA~\citet{cheng2025navila} & 8.78 & 46.8 & 34.3 & 28.2 \\
AwareVLN~\citet{guo2026awarevln} & 7.15 & 51.0 & 39.8 & 36.0 \\
\midrule
\textbf{GroundingVLN (Ours)} & \textbf{5.21} & \textbf{65.8} & \textbf{59.9} & \textbf{48.2} \\
\bottomrule
\end{tabular}
\endgroup
\end{table}

\fi

\subsection{Main Results}
\label{sec:main-results}

\paragraph{Results on VLN-CE benchmark.}

As shown in Table~\ref{tab:main-results}, GroundingVLN achieves the best results across all reported metrics on R2R-CE and RxR-CE, with absolute SR gains of 3.5\% and 5.8\% over ABot-N0, respectively. It also surpasses ETPNav, the strongest waypoint-based baseline, and pixel-goal-based DualVLN. This advantage persists against DualVLN (System 2) + SPF, which uses Habitat's Shortest Path Follower for idealized execution of System 2's predicted pixel goals. Together, these findings suggest that visually grounded reasoning and spatial goals aligned with semantic progress and execution constraints support more reliable navigation. Notably, GroundingVLN achieves these results with only 188K training samples, approximately 0.9\% of ABot-N0's total reported training data, demonstrating strong sample efficiency.

\paragraph{Cross-Dataset Generalization.}
\label{sec:cross-dataset-generalization}

\ifdefined\arxivEarlyTables\else
\begin{table}[htbp]
\centering
\caption{R2R-to-RxR transfer on the RxR-CE \texttt{val\_unseen} split without using RxR-CE training data.}
\label{tab:cross-dataset-generalization}
\begingroup
\small
\setlength{\tabcolsep}{7pt}
\renewcommand{\arraystretch}{1.05}
\begin{tabular}{@{}lcccc@{}}
\toprule
Method & NE$\downarrow$ & OS$\uparrow$ & SR$\uparrow$ & SPL$\uparrow$ \\
\midrule
NaVid~\citet{zhang2024navid} & 8.41 & 34.5 & 23.8 & 21.2 \\
MonoDream~\citet{wang2026monodream} & 8.57 & 35.9 & 25.1 & 21.6 \\
Progress-Think~\citet{wang2025progressthink} & 8.30 & 38.3 & 27.5 & 22.7 \\
NaVILA~\citet{cheng2025navila} & 8.78 & 46.8 & 34.3 & 28.2 \\
AwareVLN~\citet{guo2026awarevln} & 7.15 & 51.0 & 39.8 & 36.0 \\
\midrule
\textbf{GroundingVLN (Ours)} & \textbf{5.21} & \textbf{65.8} & \textbf{59.9} & \textbf{48.2} \\
\bottomrule
\end{tabular}
\endgroup
\end{table}

\fi

As shown in Table~\ref{tab:cross-dataset-generalization}, GroundingVLN substantially outperforms all baselines in direct R2R-to-RxR transfer without any RxR-CE training data. It achieves 59.9\% SR and 48.2\% SPL, exceeding the strongest baseline, AwareVLN, by absolute margins of 20.1\% and 12.2\%, respectively. These pronounced gains suggest that explicit subtask progress, visually grounded evidence, and the semantic pixel-goal interface provide a highly transferable navigation representation rather than overfitting to the instruction style and trajectories of R2R.

\ifdefined\arxivEarlyTables
\suppressfloats[t]
\begin{table}[!htbp]
\centering
\caption{Ablation of key components on the R2R-CE \texttt{val\_unseen} split.}
\label{tab:key-component-ablations}
\begingroup
\small
\setlength{\tabcolsep}{9pt}
\renewcommand{\arraystretch}{1.05}
\begin{tabular}{@{}lcccc@{}}
\toprule
Method & NE$\downarrow$ & OS$\uparrow$ & SR$\uparrow$ & SPL$\uparrow$ \\
\midrule
Full model & \textbf{3.66} & 74.8 & \textbf{69.9} & \textbf{64.1} \\
w/o \texttt{<obj>} and \texttt{<prev\_obj>} & 3.94 & 68.8 & 66.1 & 62.8 \\
w/o temporal alignment of grounding & 4.48 & 67.8 & 62.2 & 58.1 \\
w/o GEAR & 4.65 & 66.0 & 57.2 & 49.5 \\
w/o Subtask Decomposition & 4.05 & \textbf{74.9} & 68.1 & 62.2 \\
w/ 2D pixel-distance reward & 4.07 & 73.2 & 66.2 & 60.7 \\
\bottomrule
\end{tabular}
\endgroup
\end{table}

\begin{figure}[t]
\centering
\includegraphics[width=\linewidth]{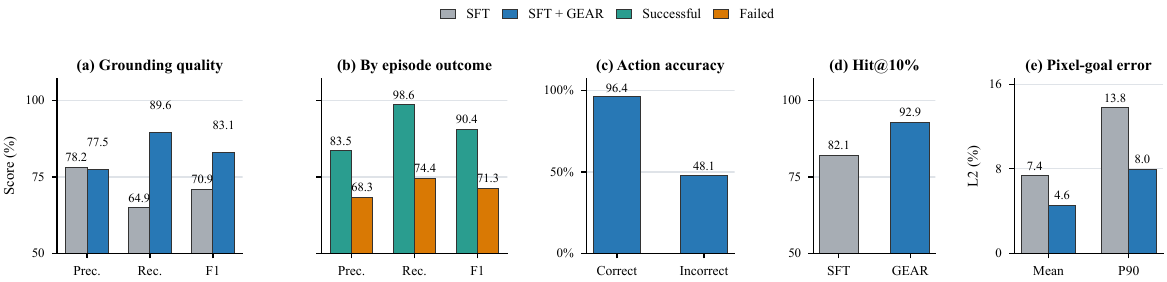}
\caption{Grounding accuracy and pixel-goal prediction. (a) Grounding quality under SFT and SFT+GEAR. (b) Grounding by episode outcome. (c) Action accuracy conditioned on grounding correctness. (d) The fraction of goals with normalized L2 error $\leq 10\%$. (e) Localization error.}
\label{fig:grounding-and-pixel-goal-quality}
\end{figure}

\begin{figure}[t]
\centering
\includegraphics[width=\linewidth]{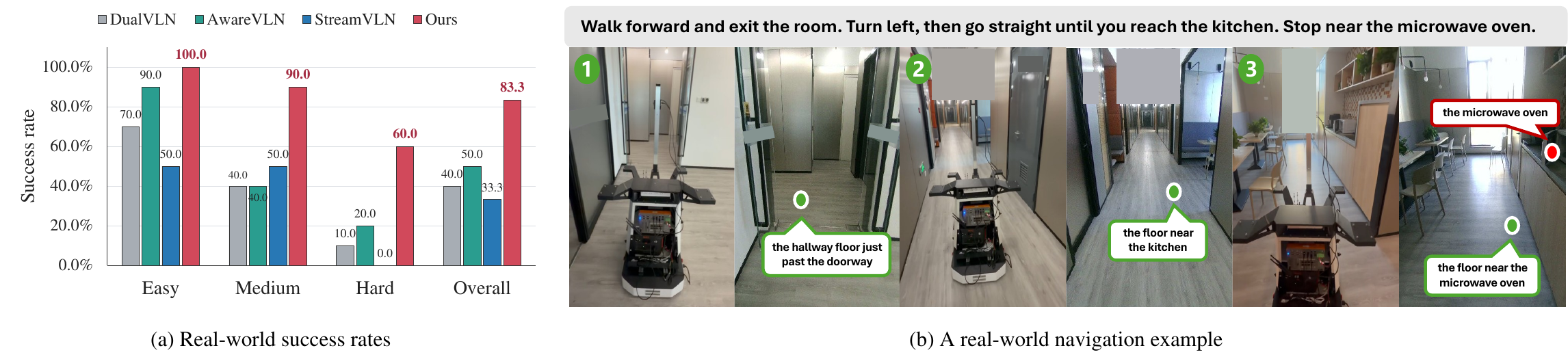}
\caption{Real-World Performance Analysis. (a) Success rates over 10 episodes per difficulty level. (b) Visualization of a real-world navigation example.}
\label{fig:real-world-results}
\end{figure}

\fi

\subsection{Ablation Study}
\label{sec:ablation-study}

\paragraph{Key Component Ablations.}
\label{sec:key-component-ablations}

\ifdefined\arxivEarlyTables\else
\begin{table}[!htbp]
\centering
\caption{Ablation of key components on the R2R-CE \texttt{val\_unseen} split.}
\label{tab:key-component-ablations}
\begingroup
\small
\setlength{\tabcolsep}{9pt}
\renewcommand{\arraystretch}{1.05}
\begin{tabular}{@{}lcccc@{}}
\toprule
Method & NE$\downarrow$ & OS$\uparrow$ & SR$\uparrow$ & SPL$\uparrow$ \\
\midrule
Full model & \textbf{3.66} & 74.8 & \textbf{69.9} & \textbf{64.1} \\
w/o \texttt{<obj>} and \texttt{<prev\_obj>} & 3.94 & 68.8 & 66.1 & 62.8 \\
w/o temporal alignment of grounding & 4.48 & 67.8 & 62.2 & 58.1 \\
w/o GEAR & 4.65 & 66.0 & 57.2 & 49.5 \\
w/o Subtask Decomposition & 4.05 & \textbf{74.9} & 68.1 & 62.2 \\
w/ 2D pixel-distance reward & 4.07 & 73.2 & 66.2 & 60.7 \\
\bottomrule
\end{tabular}
\endgroup
\end{table}

\fi

As shown in Table~\ref{tab:key-component-ablations}, all components contribute to navigation success. Removing GEAR causes the largest absolute SR drop (12.7\%), while replacing its execution-aware reward map with 2D pixel-distance supervision also degrades performance, highlighting the importance of aligning optimization with physical execution. Removing \texttt{<obj>} and \texttt{<prev\_obj>} lowers SR by 3.8 percentage points, supporting the role of grounded reasoning in linking navigation decisions to visual evidence. Disabling temporal alignment of grounding during data annotation and retraining on the resulting annotations causes a larger drop, suggesting that inconsistent cross-view evidence can mislead reasoning. Subtask decomposition further improves success and path efficiency by organizing long instructions into coherent stages.

\paragraph{Grounding Accuracy Analysis.}
\label{sec:grounding-quality-analysis}

To assess GEAR's effect on grounding quality and its relationship with navigation accuracy, we manually annotate 100 episodes from R2R-CE \texttt{val\_unseen}. Figure~\ref{fig:grounding-and-pixel-goal-quality}(a) compares SFT with SFT+GEAR, showing improved grounding recall and F1 after GEAR. Figure~\ref{fig:grounding-and-pixel-goal-quality}(b) compares grounding accuracy between successful and failed episodes, showing higher grounding accuracy in successful episodes. Figure~\ref{fig:grounding-and-pixel-goal-quality}(c) examines the association between grounding correctness in reasoning and action correctness across all decision steps. Correct grounding is associated with nearly twice the action accuracy, linking reliable visual evidence to better decisions. Higher Hit@10\% and lower mean and tail localization errors in Figure~\ref{fig:grounding-and-pixel-goal-quality}(d,e) further indicate that GEAR improves spatial goal precision alongside grounding accuracy. These results show that GEAR substantially strengthens grounding during both reasoning and decision making, yielding a large improvement in final navigation success.

\ifdefined\arxivEarlyTables\else
\begin{figure}[t]
\centering
\includegraphics[width=\linewidth]{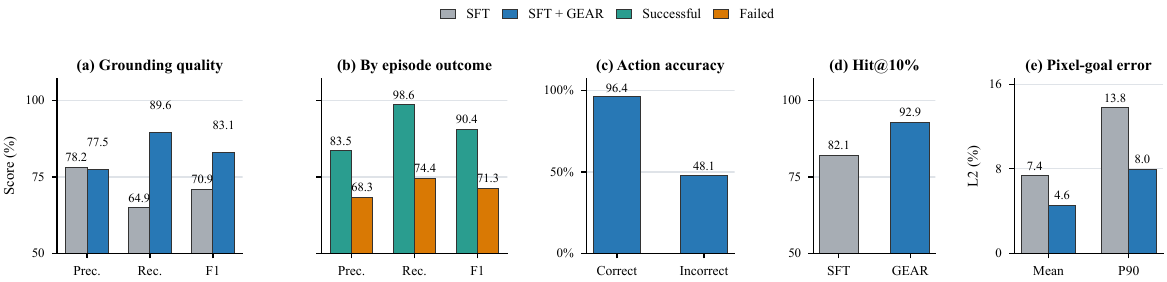}
\caption{Grounding accuracy and pixel-goal prediction. (a) Grounding quality under SFT and SFT+GEAR. (b) Grounding by episode outcome. (c) Action accuracy conditioned on grounding correctness. (d) The fraction of goals with normalized L2 error $\leq 10\%$. (e) Localization error.}
\label{fig:grounding-and-pixel-goal-quality}
\end{figure}

\fi

\ifdefined\arxivEarlyTables\else
\begin{figure}[t]
\centering
\includegraphics[width=\linewidth]{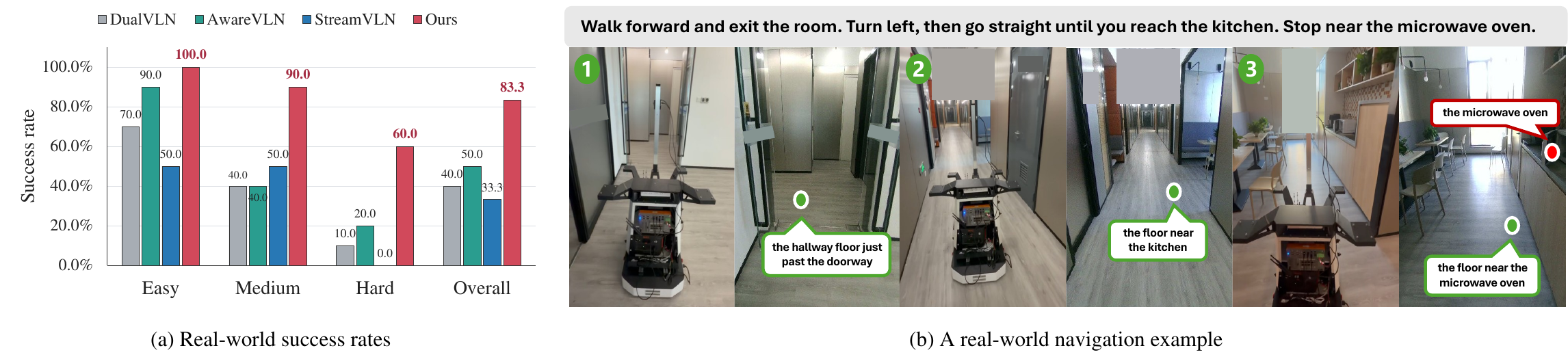}
\caption{Real-World Performance Analysis. (a) Success rates over 10 episodes per difficulty level. (b) Visualization of a real-world navigation example.}
\label{fig:real-world-results}
\end{figure}

\fi

\subsection{Real-World Deployment}
\label{sec:real-world-deployment}

We conduct real-world experiments on an AgileX TRACER 2.0 wheeled robot equipped with an Insta360 X5 and an Intel RealSense D435. The camera setup uses a mounting height of 1.2\,m and a downward tilt of $30^\circ$. GroundingVLN inference runs on a single NVIDIA GeForce RTX 4090 GPU. 
As shown in Figure~\ref{fig:real-world-results}(a), GroundingVLN outperforms all three baselines across every difficulty level. Its advantage is especially pronounced on medium and hard tasks. GroundingVLN is substantially more robust in translating compositional language instructions into temporally consistent navigation behaviors, particularly when successful execution requires maintaining intermediate goals, respecting spatial constraints, and performing long-horizon action transitions.

\endgroup
\section{Conclusion}
\label{sec:conclusion}

We introduced \textbf{GroundingVLN}, a grounding-centric framework that connects high-level reasoning and low-level execution through visual grounding. It anchors task-relevant evidence during structured reasoning and translates semantic progress into executable pixel goals. \textbf{GroundingCOTVLN-188K} provides temporally aligned grounded supervision, while \textbf{GEAR} aligns reasoning and spatial decisions with downstream execution. Experiments demonstrate state-of-the-art performance, strong data efficiency, and cross-dataset generalization, showing that visual grounding is an effective interface for adapting pretrained VLMs to embodied navigation.

\bibliography{iclr2027_conference}
\bibliographystyle{iclr2027_conference}

\appendix
\section{Appendix}
\label{sec:appendix}

\subsection{Training-Data Efficiency}
\label{sec:appendix-data-efficiency}

\begin{figure}[H]
\centering
\includegraphics[width=0.88\linewidth]{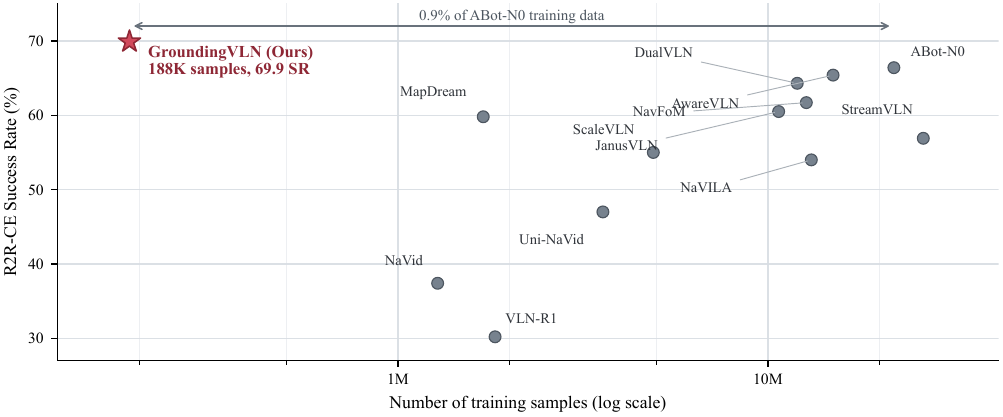}
\caption{Training-data efficiency on R2R-CE \texttt{val\_unseen}. The horizontal axis shows each paper's reported training-example count in its native unit on a logarithmic scale; methods without a reported non-zero count are omitted. GroundingVLN achieves the highest SR among the plotted methods with 188K examples, equivalent to 0.9\% of ABot-N0's reported total.}
\label{fig:r2r-data-efficiency}
\end{figure}

We use each paper's reported training count without converting between units or double-counting reused data. Across methods, a reported example may be a trajectory, a clip, or a decision step; a trajectory or clip can contain multiple steps. Our 188K examples are step-level VLM input--output pairs: each pairs the current observation and up to three context images and preceding reasoning turns with one grounded high-level decision. They come from 21K R2R and RxR trajectories and include the 16K GEAR subset. ABot-N0~\citep{chu2026abotn0} reports 16.9M training trajectories across five navigation paradigms and 5.0M reasoning examples, totaling 21.9M. Its reasoning data include navigability labels, navigation CoTs generated by Qwen-VL-Max, Gemini-3 Pro, and Qwen3-VL, trajectory-verified Object-Goal reasoning, POI-grounding VQA, and public VQA/EQA data.

Figure~\ref{fig:r2r-data-efficiency} shows that GroundingVLN reaches the highest R2R-CE \texttt{val\_unseen} SR among the plotted methods with 188K decision-step examples, or $0.86\%$ (rounded to 0.9\%) of ABot-N0's reported total. Since a training trajectory can contain multiple decision steps, a step-level count for ABot-N0 could be larger and the corresponding proportion smaller than 0.9\%.

\subsection{Comparison without Additional Training Data}
\label{sec:appendix-benchmark-only-comparison}

\begin{table}[H]
\centering
\caption{Comparison with the benchmark-only and full-data variants of StreamVLN and JanusVLN on the \texttt{val\_unseen} splits. Benchmark-only variants are trained using R2R and RxR data without additional navigation corpora; their exact sample counts are not reported.}
\label{tab:benchmark-only-data-efficiency}
\begingroup
\small
\setlength{\tabcolsep}{3.6pt}
\renewcommand{\arraystretch}{1.06}
\resizebox{\linewidth}{!}{%
\begin{tabular}{@{}llc@{\hspace{7pt}}ccc@{\hspace{7pt}}cccc@{}}
\toprule
& & & \multicolumn{3}{c}{R2R-CE Val-Unseen} & \multicolumn{4}{c}{RxR-CE Val-Unseen} \\
\cmidrule(lr){4-6}\cmidrule(lr){7-10}
Method & Training corpus & Samples & NE$\downarrow$ & SR$\uparrow$ & SPL$\uparrow$ & NE$\downarrow$ & SR$\uparrow$ & SPL$\uparrow$ & nDTW$\uparrow$ \\
\midrule
StreamVLN~\citep{wei2026streamvln} & R2R+RxR only & -- & 5.98 & 45.6 & 42.3 & -- & -- & -- & -- \\
StreamVLN~\citep{wei2026streamvln} & Full corpus & $\sim$26.3M & 4.98 & 56.9 & 51.9 & 6.22 & 52.9 & 46.0 & 61.9 \\
\addlinespace[2pt]
JanusVLN Base~\citep{zeng2026janusvln} & R2R+RxR only & -- & 5.17 & 52.8 & 49.2 & 6.46 & 51.4 & 44.3 & 59.1 \\
JanusVLN~\citep{zeng2026janusvln} & Full corpus & 10.69M & 4.78 & 60.5 & 56.8 & 6.06 & 56.2 & 47.5 & 62.1 \\
\midrule
\textbf{GroundingVLN (Ours)} & R2R+RxR only & \textbf{188K} & \textbf{3.66} & \textbf{69.9} & \textbf{64.1} & \textbf{3.54} & \textbf{75.1} & \textbf{62.0} & \textbf{75.3} \\
\bottomrule
\end{tabular}%
}
\endgroup
\end{table}

Table~\ref{tab:benchmark-only-data-efficiency} further controls for the use of external navigation corpora. With only 188K R2R and RxR samples, GroundingVLN achieves absolute SR gains of 24.3\% and 17.1\% over the benchmark-only variants of StreamVLN and JanusVLN on R2R-CE, respectively, and 23.7\% over JanusVLN Base on RxR-CE. More importantly, it remains substantially stronger than their full-data variants: GroundingVLN improves R2R-CE SR by 13.0\% over StreamVLN and 9.4\% over JanusVLN, and improves RxR-CE SR by 18.9\% over JanusVLN, while using only 0.7\% and 1.8\% of their respective training samples. These results show that the gains arise from more effective grounded supervision and execution alignment rather than from scaling the training corpus.

\subsection{General-Capability Retention}
\label{sec:appendix-general-capability}

To assess whether navigation tuning preserves the pretrained model's general multimodal capabilities, we evaluate zero-shot performance on MMStar~\citep{chen2024mmstar} and MVBench~\citep{li2024mvbench} under the same deterministic protocol.
\begin{table}[H]
\centering
\caption{Zero-shot accuracy (\%) on general multimodal benchmarks after navigation tuning.}
\label{tab:general-capability-retention}
\begingroup
\small
\setlength{\tabcolsep}{7pt}
\renewcommand{\arraystretch}{1.05}
\begin{tabular}{@{}lcc@{}}
\toprule
Model & MMStar & MVBench \\
\midrule
Qwen3.5-4B & \textbf{63.67} & \textbf{50.93} \\
GroundingVLN & 57.20 & 45.55 \\
AwareVLN~\citep{guo2026awarevln} & 0.07 & 0.00 \\
DualVLN System~2~\citep{wei2026dualvln} & 0.00 & 0.00 \\
\bottomrule
\end{tabular}
\endgroup
\end{table}

As shown in Table~\ref{tab:general-capability-retention}, GroundingVLN largely preserves general multimodal competence without mixing general-purpose data into navigation training, unlike the severe specialization often observed in end-to-end VLN agents. Compared with the original Qwen3.5-4B, it shows absolute accuracy drops of only 6.47\% on MMStar and 5.38\% on MVBench. Moreover, navigation tuning yields absolute accuracy gains of 27.5\% on Object Shuffle, 5.0\% on Scene Transition, 4.5\% on Counterfactual Inference, 3.5\% on State Change, and 3.0\% on Egocentric Navigation. These gains are concentrated in spatial-change tracking, temporal transition understanding, and egocentric motion reasoning, suggesting that navigation data strengthen capabilities directly relevant to embodied interaction.

For comparison, we evaluate AwareVLN and DualVLN System~2, two VLN models likewise adapted from general-purpose VLM backbones. Despite using prompts adapted to the VQA tasks, both models remain largely unable to provide the required answer choices: AwareVLN continues to emit its navigation protocol, while DualVLN System~2 returns directional actions or \texttt{STOP}. This collapse toward a navigation-specific output space is a clear manifestation of catastrophic forgetting under task-specific fine-tuning. Taken together, GroundingVLN combines state-of-the-art VLN performance with strong retention of its pretrained general capabilities, highlighting its potential as a general-purpose embodied foundation model.

\subsection{Inference Efficiency}
\label{sec:appendix-efficiency}

\begin{table}[H]
\centering
\caption{Average inference time per navigation episode. Lower is better.}
\label{tab:inference-efficiency}
\begingroup
\small
\setlength{\tabcolsep}{12pt}
\renewcommand{\arraystretch}{1.05}
\begin{tabular}{@{}lc@{}}
\toprule
Method & Time (s/episode)$\downarrow$ \\
\midrule
NaVILA~\citet{cheng2025navila} & 56.84 \\
Aux-Think~\citet{wang2025auxthink} & 58.04 \\
Progress-Think~\citet{wang2025progressthink} & \textbf{36.60} \\
ActiveVLN~\citet{zhang2026activevln} & 106.90 \\
\textbf{GroundingVLN (Ours)} & 37.41 \\
\bottomrule
\end{tabular}
\endgroup
\end{table}

We measure inference time using a single NVIDIA RTX 4090 GPU. As shown in Table~\ref{tab:inference-efficiency}, GroundingVLN requires 37.41 seconds per episode, comparable to the fastest Progress-Think result of 36.60 seconds, while reducing latency by 34.2\%, 35.5\%, and 65.0\% relative to NaVILA, Aux-Think, and ActiveVLN, respectively. GroundingVLN invokes the VLM only 9.26 times per episode on average, corresponding to 33.25\% of navigation steps. The low-level planner handles execution between these sparse high-level decisions, avoiding repeated VLM inference at every primitive action while retaining structured reasoning at key decision points.

\subsection{Robustness to Depth Noise}
\label{sec:appendix-depth-noise}

To assess sensitivity to imperfect depth sensing, we evaluate GroundingVLN on R2R-CE \texttt{val\_unseen} with a disparity-domain noise model configured using the D435 stereo baseline and subpixel error characteristics documented by Intel RealSense. For each finite ground-truth depth $Z$ satisfying $0.17<Z<5\,\mathrm{m}$, we compute
\begin{equation}
\begin{aligned}
d &= \frac{f_xB}{Z},
& \tilde d &= d+\epsilon_d, \qquad \epsilon_d\sim\mathcal{N}(0,\sigma_d^2),\\
\hat d &= q\,\operatorname{round}(\tilde d/q),
& \hat Z &= \frac{f_xB}{\hat d},
\end{aligned}
\label{eq:stereo-depth-noise}
\end{equation}
where $f_x$ is each camera's original horizontal focal length in pixels, $B=0.05\,\mathrm{m}$ is the stereo baseline, $\sigma_d=0.08$ pixels, and $q=1/32$ pixel is the disparity quantization step. Invalid input depths, non-positive disparities, and reconstructed depths that are non-finite or outside the same valid range are set to zero. Out-of-range depths are not clamped to $5\,\mathrm{m}$, avoiding artificial distant surfaces. Ignoring quantization and range filtering, the first-order depth uncertainty grows quadratically with distance, $\sigma_Z\approx Z^2\sigma_d/(f_xB)$. For the front camera with $f_x\approx277.13$ pixels, this corresponds to approximately $0.58$, $5.20$, and $14.43$\,cm at $1$, $3$, and near $5$\,m, respectively.

\begin{table}[H]
\centering
\caption{Depth-noise robustness on R2R-CE \texttt{val\_unseen}. NE is in meters; OS, SR, and SPL are percentages.}
\label{tab:depth-noise-robustness}
\begingroup
\small
\setlength{\tabcolsep}{7pt}
\renewcommand{\arraystretch}{1.05}
\begin{tabular}{@{}lcccc@{}}
\toprule
Depth input & NE$\downarrow$ & OS$\uparrow$ & SR$\uparrow$ & SPL$\uparrow$ \\
\midrule
Ground-truth depth & 3.66 & 74.8 & 69.9 & 64.1 \\
D435-like noisy depth & 4.14 & 71.2 & 65.5 & 59.7 \\
\bottomrule
\end{tabular}
\endgroup
\end{table}

As shown in Table~\ref{tab:depth-noise-robustness}, GroundingVLN retains an SR of 65.5\% and an SPL of 59.7\% under noisy depth, with an absolute decrease of 4.4\% in both metrics relative to ground-truth depth. The moderate degradation suggests that its pixel-goal execution pipeline tolerates the tested distance-dependent measurement errors, although reliable depth remains important for geometric planning.

\subsection{Qualitative Results}
\label{sec:appendix-qualitative-results}

\begin{figure}[!t]
    \centering
    \includegraphics[width=\textwidth]{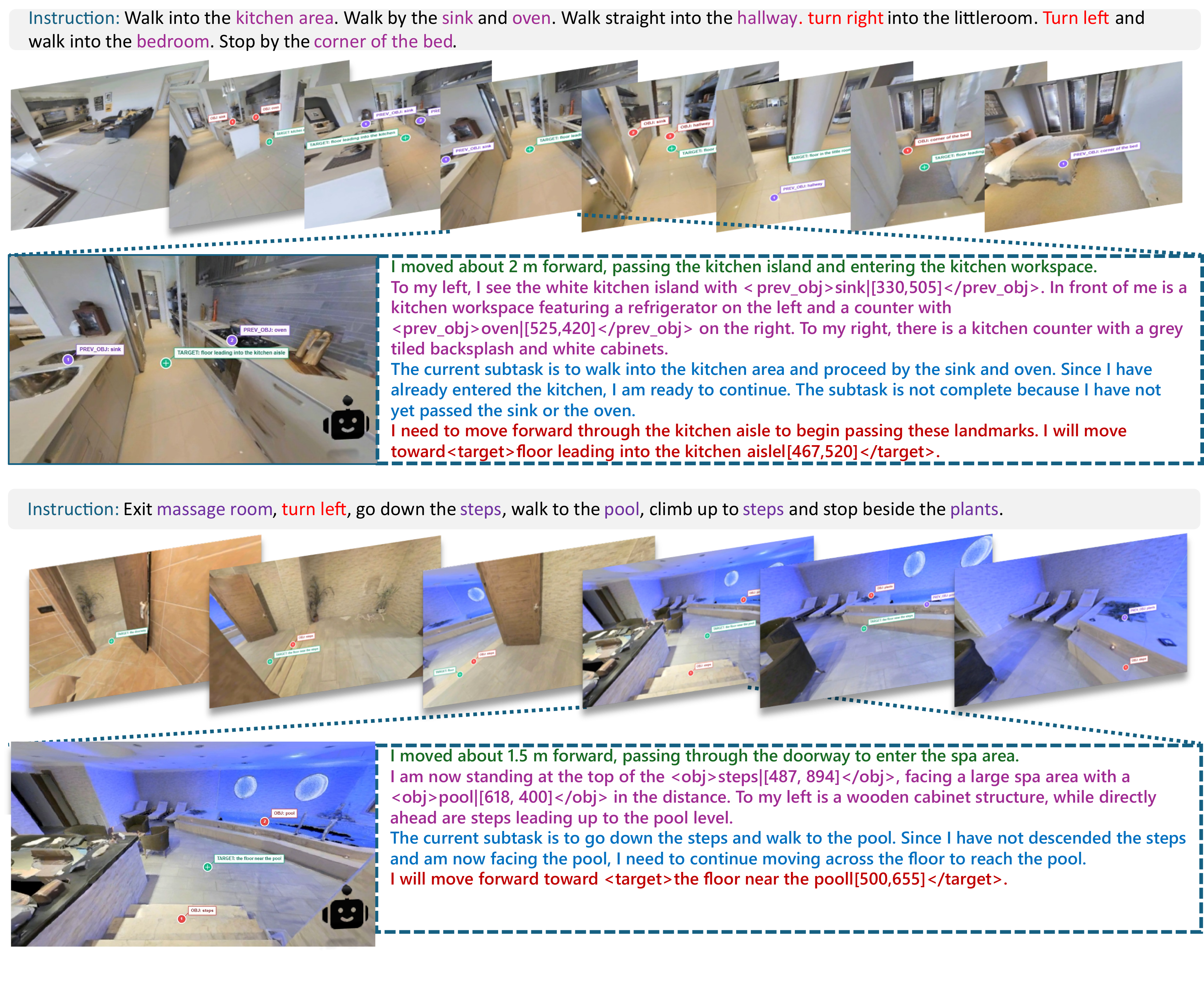}
    \caption{Two qualitative examples of GroundingVLN. The upper example follows a route from the kitchen to the bedroom, while the lower example navigates from a massage room toward a pool and nearby plants. Each example shows the instruction, representative trajectory observations, and an enlarged decision step with grounded reasoning. Newly grounded evidence, recalled evidence, and pixel goals are marked by \texttt{<obj>}, \texttt{<prev\_obj>}, and \texttt{<target>}, respectively.}
    \label{fig:qualitative-results}
\end{figure}

Figure~\ref{fig:qualitative-results} illustrates how grounded reasoning guides navigation in two distinct environments. In the kitchen-to-bedroom example, the model recalls the sink and oven as historical evidence despite viewpoint changes. At the highlighted step, it distinguishes entering the kitchen from passing these landmarks and selects a pixel goal further along the aisle. In the massage-room-to-pool example, the model grounds the steps and pool, recognizes that it has entered the spa area but remains at the top of the steps, and selects a floor target toward the pool while identifying descent and approach as unfinished parts of the subtask. Both examples connect visual evidence and completed motion to the remaining subtask, illustrating how temporal grounding and semantic progress guide local goal selection along longer routes.

\subsection{GroundingCOTVLN-188K Data-Engine Details}
\label{sec:appendix-data-engine}

This section details the automatic data engine introduced in Section~\ref{sec:sft-data-engine}. The pipeline first replays reference trajectories to collect temporally ordered anchors. It then annotates each anchor with its current subtask, progress state, visual observation, and instruction-relevant entities. Finally, it aligns grounded evidence across time, derives action and pixel-goal labels from the executed geometry, and filters invalid examples before export.

\paragraph{Replay and anchor construction.}
For each episode, the replay stage executes the reference route with the same low-level planner used by the navigation system and caches the \emph{executed} trajectory rather than teleporting between reference poses. The cache additionally retains camera intrinsics, poses, actions, waypoint indices, and the complete left--front--right frame stream required by temporal tracking. Annotation then runs in a strict cache-only mode, which decouples simulator throughput from VLM and SAM~3 inference. 

Before replay, dense routes containing more than ten points or an adjacent pair closer than $1\,$m are resampled at approximately uniform arc-length intervals. The endpoints and the strongest route turns above $45^\circ$ are explicitly retained. If merging is unnecessary, only segments longer than $4\,$m are interpolated, with a maximum spacing of $3\,$m. We record arrival anchors at reference points, paired pre-/post-turn anchors for executed in-place rotations of at least $60^\circ$, and mid-distance anchors after approximately $3\,$m of travel. Candidate anchors triggered by $45^\circ$ accumulated rotation or 15 low-level steps are used to fill retained trajectory gaps exceeding $4\,$m. Near-static anchors immediately preceding an arrival or turn are removed. To diversify termination states, the final stopping radius is sampled independently for every episode from $[0,1]\,$m.

\paragraph{Subtask, progress, and observation annotation.}
The teacher first divides the instruction into ordered, exact instruction substrings and associates each subtask with a checklist of two or three landmarks or turns. The decomposition is rejected and retried when a subtask is not an instruction substring, a checklist has an invalid structure, or a turn is unsupported by its subtask. A progress annotator then receives at most 20 sampled front-view frames, the executed action trace, waypoint events, and sparse room hints, and aligns every subtask with a contiguous waypoint interval. Turn boundaries are subsequently repaired using the measured signed rotations, preventing visually plausible but temporally incorrect progress labels.

Observation generation is performed before CoT synthesis. For the current anchor, the teacher sees the unannotated stitched panorama; from the second anchor onward, it also receives the preceding successful-anchor image and a coordinate-free version of the preceding observation. Task-relevant object candidates are selected in batches of at most ten anchors. The front view is processed first, and a full left--front--right pass is invoked only when instruction landmarks remain uncovered. Directed grounding is then performed independently within each view, prioritizing the front panel. Panel-local coordinates are normalized to $[0,1000]^2$ and converted to the common stitched-image coordinate system. Outputs referring to oracle overlays, reference points, target annotations, or other generation artifacts are rejected and regenerated.

\paragraph{Cross-time grounding alignment.}
Each newly accepted \texttt{<obj>} coordinate is paired with depth and back-projected through the anchor camera pose to create a persistent world-space entity record. Initial estimates farther than $5\,$m from the agent are discarded. The corresponding box prompt initializes SAM~3 on the complete per-step replay stream; when only a point is available, the engine constructs a small point-centered box. We use a mask-presence threshold of $0.4$ and a minimum mask area of 64 pixels. The representative point is chosen near the mask centroid from pixels with valid depth and is then back-projected to refresh the world position.

Temporal identity is accepted only when the refreshed 3D point is within $2\,$m of the previous valid estimate. Tracks with the same normalized label and mask IoU above $0.5$ are merged, while broad regions such as \emph{floor}, \emph{room}, and \emph{hallway} are excluded from instance tracking. A SAM~3 miss or a mask without usable depth does not overwrite a valid world position; that position is retained for up to five CoT anchors. Consequently, an entity can leave the image and later be reprojected with the same identity. Such reappearances are encoded as \texttt{<prev\_obj>}, whereas only newly established evidence uses \texttt{<obj>}.

\paragraph{Geometry-owned action and pixel-goal labels.}
After observation grounding, a deterministic assembler derives progress, actions, and pixel goals from the executed trajectory. For \texttt{MOVE}, the trajectory is densified to at most $0.25\,$m spacing, and the target is the farthest continuously visible executed point before the next retained anchor and within the current semantic segment. Projection searches for a ground-contact pixel and checks depth agreement, occlusion, front-view membership, and traversability; a \texttt{MOVE} anchor is dropped when no genuine target is visible. A \texttt{TURN} target is optional and is kept only when its view is compatible with the rotation direction and the teacher confirms a visible semantic cue. \texttt{STOP} is emitted only after the final subtask is complete and the executed pose lies within the replay terminal tolerance. The teacher verbalizes these facts and supplies a concise target label, but post-processing overwrites its action and coordinate fields with the deterministic labels. It also enforces one matching \texttt{<target>} tag whenever a pixel goal is present.

\paragraph{Export and quality control.}
Every anchor exports a raw panorama, a rendered visual-memory image, and a variant without historical key-object markings. The current response is added to visual and textual memory only \emph{after} its input has been rendered, preventing its supervised labels from leaking into the same sample. Each example contains at most three prior successful reasoning turns and three between-anchor context images.

Before export, the engine requires a successful teacher status, a non-empty structured \texttt{<think>} trace, valid grounding markup, action--reasoning agreement, and a visible target for \texttt{MOVE}; it also rejects annotation leakage and malformed turn supervision. An offline audit revalidates the mixed \texttt{<think>}--JSON schema, agreement between parsed and serialized targets, message--image chains, coordinates, and every assistant response retained in the history. It further checks decomposition substring/checklist constraints and removes all remaining samples from a navigation episode if no valid \texttt{STOP} survives row-level filtering.

\begin{table}[H]
\centering
\caption{Key implementation settings of the GroundingCOTVLN-188K data engine.}
\label{tab:data-engine-settings}
\begingroup
\small
\setlength{\tabcolsep}{4pt}
\renewcommand{\arraystretch}{1.02}
\begin{tabular}{@{}p{0.59\columnwidth}p{0.31\columnwidth}@{}}
\toprule
Setting & Value \\
\midrule
Route target budget / minimum spacing & $10$ / $1\,$m \\
Turn preservation / turn-anchor threshold & $45^\circ$ / $60^\circ$ \\
Anchor stride / gap-insertion threshold & $3\,$m / $4\,$m \\
Progress frames / history turns / context frames & $20$ / $3$ / $3$ \\
Key-object annotation batch size & $10$ anchors \\
SAM~3 presence / minimum mask area & $0.4$ / $64$ px \\
World-point range / identity distance & $5\,$m / $2\,$m \\
Duplicate-mask IoU / state retention & $0.5$ / $5$ anchors \\
Executed-trajectory target spacing & $0.25\,$m \\
Sampled stop threshold / tolerance floor & $[0,1]\,$m / $0.35\,$m \\
\bottomrule
\end{tabular}
\endgroup
\end{table}

\subsection{Coordinate-Aware Supervised Fine-Tuning}
\label{sec:appendix-sft-coordinate}

In addition to the token-weighted autoregressive cross-entropy $\mathcal{L}_{\mathrm{wCE}}$, SFT uses a differentiable coordinate objective. At each coordinate digit position $k$, we restrict the logits to digit tokens and compute the expected digit $\hat d_k=\sum_{d=0}^{9}d\,p_k(d)$. The expected coordinate is then reconstructed from its decimal place values $v_k$ as $\hat c=\sum_k v_k\hat d_k$. For all coordinates $c$ appearing in \texttt{<obj>}, \texttt{<prev\_obj>}, \texttt{<target>}, and the action JSON, we optimize
\begin{equation}
    \mathcal{L}_{\mathrm{SFT}}
    =\mathcal{L}_{\mathrm{wCE}}
    +\lambda_{\mathrm{coord}}\frac{1}{|\mathcal{C}|}
    \sum_{c\in\mathcal{C}}
    \operatorname{SmoothL1}_{\delta}\!\left(
        \frac{\hat c}{s_{\mathrm{coord}}},
        \frac{c}{s_{\mathrm{coord}}}
    \right).
    \label{eq:sft-coordinate-loss}
\end{equation}

\subsection{Detailed GEAR Reward Definitions}
\label{sec:appendix-reward-definitions}

We describe the reward functions used in our implementation. Let $\bar a_t$, $\bar d_t$, $\bar c_t$, and $\bar p_t$ denote the annotated action, turn direction, current subtask, and pixel goal, respectively, and let the corresponding predictions be denoted with hats. For pixel-goal supervision, our implementation uses the execution-aware target maps described below.

\paragraph{Strict output validity.}
Before assigning a semantic reward, we require exactly one non-empty \texttt{<think>} block followed by one action JSON object and no trailing text. The JSON must contain a non-empty current subtask and an action-specific schema. \texttt{MOVE} requires one target with an integer coordinate in $[0,1000]^2$, \texttt{TURN} requires a valid direction and may contain one target, and \texttt{STOP} must not contain a target. A target-bearing decision must also contain exactly one matching \texttt{<target>} tag in the reasoning. Malformed, truncated, schema-inconsistent, or coordinate-inconsistent outputs receive
\begin{equation}
R_{\mathrm{invalid}}=-1.
\label{eq:gear-invalid-reward}
\end{equation}

\paragraph{Action and subtask rewards.}
For valid outputs, the action reward is the following deterministic matrix:
\begin{equation}
R_{\mathrm{act}}=
\begin{cases}
 1, & \hat a_t=\bar a_t\ \text{and, for \texttt{TURN}, }\hat d_t=\bar d_t,\\
-0.5, & \bar a_t=\hat a_t=\texttt{TURN},\ \hat d_t\ne\bar d_t,\\
-0.75, & \bar a_t=\texttt{TURN},\ \hat a_t=\texttt{MOVE},\ \hat p_t\in\mathcal{Q}_{\bar d_t},\\
-0.9, & (\bar a_t,\hat a_t)\in\{(\texttt{MOVE},\texttt{TURN}),
(\texttt{TURN},\texttt{MOVE})\},\\
-1, & \text{otherwise}.
\end{cases}
\label{eq:gear-action-reward}
\end{equation}
Here $\mathcal{Q}_{\bar d_t}$ is the outer quarter of the front panel consistent with a required left or right turn. The $-0.9$ case for a predicted \texttt{MOVE} applies when its target is outside this partially aligned region. Premature \texttt{STOP} and any wrong action for a gold \texttt{STOP} receive $-1$. For subtask consistency, we normalize Unicode, lowercase text, collapse whitespace, and remove trailing punctuation, denoted by $\nu(\cdot)$, and use
\begin{equation}
R_{\mathrm{task}}=2\,\mathbf{1}\!\left[\nu(\hat c_t)=\nu(\bar c_t)\right]-1.
\label{eq:gear-task-reward}
\end{equation}

\paragraph{Execution-aware target quality.}
For a normalized pixel $p$, depth unprojection gives a 3D endpoint $P_t(p)$. From the stitched RGB-D observation, we reconstruct a connected local support surface that preserves traversable floors, stairs, and landings while excluding occupied or unsupported regions. We set $V_{\mathrm{nav}}(p)=1$ when the endpoint is sufficiently close to this support surface, and $0$ otherwise.

Let $s_t^a$ be the agent's arc-length position on the reference route, $s(p)$ the route projection of $P_t(p)$, and $s_t^*$ the projection of the annotated endpoint $P_t^*$. We further define the executed distance $d_a(p)=\lVert P_t(p)-P_t^a\rVert_2$, forward progress $\Delta s(p)=s(p)-s_t^a$, lateral route distance $d_{\perp}(p)$, and endpoint error $d_e(p)=\lVert P_t(p)-P_t^*\rVert_2$. For a target at distance $d_t^*=\lVert P_t^*-P_t^a\rVert_2$, the near-target tightening factor is
\begin{equation}
\kappa_t=\operatorname{clip}\!\left(\frac{d_t^*}{2.0},0.33,1.0\right).
\label{eq:gear-near-target-scale}
\end{equation}
The score-validity indicator is
\begin{equation}
\begin{aligned}
V_{\mathrm{score}}(p)={}&V_{\mathrm{nav}}(p)\,
\mathbf{1}[p\in\mathcal{F}_t]\,
\mathbf{1}[d_a(p)\ge\eta_d]\\
&\cdot\mathbf{1}[\Delta s(p)\ge\eta_s]\,
\mathbf{1}[d_{\perp}(p)\le\kappa_t\eta_{\perp}]\,
\mathbf{1}[|\Delta h(p)|\le\eta_h],
\end{aligned}
\label{eq:gear-target-validity}
\end{equation}
where $\mathcal{F}_t$ is the front panel for a gold \texttt{MOVE} and the full stitched view for a target-bearing \texttt{TURN}, and $\Delta h(p)$ is the endpoint's height residual from the local support surface. The thresholds are $\eta_d=\eta_s=0.30\,\mathrm{m}$, $\eta_{\perp}=1.50\,\mathrm{m}$, and $\eta_h=0.75\,\mathrm{m}$. For a valid point, the continuous target quality is
\begin{equation}
\begin{aligned}
Q_t(p)={}&0.25\exp\!\left[-\frac{d_{\perp}(p)^2}{2(0.50\kappa_t)^2}\right]
+0.35\exp\!\left[-\frac{(s(p)-s_t^*)^2}{2(0.75)^2}\right]\\
&+0.40\exp\!\left[-\frac{d_e(p)^2}{2(1.00\kappa_t)^2}\right].
\end{aligned}
\label{eq:gear-target-quality-detailed}
\end{equation}
Invalid points have $Q_t(p)=0$.

The normalized image is discretized into a $256\times68$ map. For the set $\Omega_j$ of integer normalized coordinates assigned to map cell $j$, we store the conservative cell score
\begin{equation}
Q_{t,j}=\min_{p\in\Omega_j}Q_t(p),\qquad
V_{t,j}=\mathbf{1}\!\left[\min_{p\in\Omega_j}V_{\mathrm{score}}(p)=1\right]
\mathbf{1}[Q_{t,j}\ge0.25].
\label{eq:gear-cell-reward}
\end{equation}
A five-point center-and-corner stencil first rejects clearly invalid cells, after which every integer coordinate in each provisional positive cell is checked. Separately, the cell-level physical indicator $V^{\mathrm{nav}}_{t,j}$ is true when any stencil point has navigable support, without applying route-progress or front-panel gates; this distinguishes a physical obstacle from a walkable but off-route target. The final quality is quantized to 8 bits. We retain a target-bearing training anchor only when its annotated cell satisfies $V_{t,j^*}=1$ and $Q_{t,j^*}\ge0.80$. At reward time, the predicted target receives
\begin{equation}
R_{\mathrm{goal}}(\hat p_t)=
\begin{cases}
2Q_{t,j(\hat p_t)}-1, & V_{t,j(\hat p_t)}=1,\\
-1, & \text{otherwise}.
\end{cases}
\label{eq:gear-goal-reward}
\end{equation}
A missing target or a prediction from a different action family also receives $-1$. Because multiple target labels may be valid, $R_{\mathrm{goal}}$ does not consider exact label agreement.

\paragraph{Grounding reward.}
Teacher-VLM grounding annotations may contain occasional coordinate errors or omissions, while grounding correctness is strongly associated with decision correctness (Figure~\ref{fig:grounding-and-pixel-goal-quality}(c)). To avoid misleading supervision from imperfect annotations, we activate the auxiliary grounding bonus only when the navigation decision is also correct.

Let $\mathcal{L}_t^*$ be the set of normalized teacher grounding labels and $\widehat{\mathcal{G}}_t=\{(\hat l_k,\hat p_k)\}$ the label--coordinate pairs in predicted \texttt{<obj>} and \texttt{<prev\_obj>} tags, with label set $\widehat{\mathcal{L}}_t$. Once the required labels are present, each matched label $l$ is used as an open-vocabulary prompt to SAM 3~\citep{carion2026sam3}, which segments all of its visible instances. Let $\mathcal{M}_t(l)$ denote the union of these instance masks and define
\begin{equation}
C_t=\mathbf{1}\!\left[
\forall l\in\mathcal{L}_t^*,\ \exists(\hat l_k,\hat p_k)\in\widehat{\mathcal{G}}_t:
\hat l_k=l\ \wedge\ \hat p_k\in\mathcal{M}_t(l)
\right].
\label{eq:gear-grounding-coordinate-validity}
\end{equation}
We deliberately relax coordinate validation by accepting any point inside a visible instance mask for the matching label, rather than requiring agreement with the teacher's exact coordinates. Define $D_t=1$ when the predicted decision is correct. For a gold \texttt{MOVE}, this requires a predicted \texttt{MOVE} whose target is in the front panel and physically navigable; for \texttt{TURN} and \texttt{STOP}, it requires the correct action and, for \texttt{TURN}, the correct direction. A target-bearing \texttt{TURN} does not require an accurate target to pass $D_t$ because its target is auxiliary shaping. The grounding bonus is
\begin{equation}
B_{\mathrm{grd}}=0.10\,\mathbf{1}[D_t=1]\,
\mathbf{1}[\emptyset\ne\mathcal{L}_t^*\subseteq\widehat{\mathcal{L}}_t]\,
\mathbf{1}[C_t=1]\,
\mathbf{1}[T_t=1]\,\mathbf{1}[A_t=0],
\label{eq:gear-grounding-bonus}
\end{equation}
where $T_t=1$ for a target-free annotation, or when the predicted target is score-valid and has $Q\ge0.50$; $A_t=1$ if any predicted grounding label contains \emph{move} or \emph{turn}. A missing mask or a matched point outside all corresponding masks prevents the bonus. Extra object labels do not prevent the bonus. They are penalized only in a bad decision context:
\begin{equation}
P_{\mathrm{grd}}=0.10\,\mathbf{1}\!\left[
(R_{\mathrm{act}}<1)\ \vee\ (\bar a_t=\texttt{MOVE}\wedge \neg V_{t,j(\hat p_t)})
\right]
\mathbf{1}\!\left[\widehat{\mathcal{L}}_t\setminus\mathcal{L}_t^*\ne\emptyset\right].
\label{eq:gear-grounding-penalty}
\end{equation}
Thus, a missing label or spatially invalid coordinate forfeits the bonus without an additional negative penalty, while an unsupported extra label accompanying an incorrect action or out-of-score \texttt{MOVE} is penalized.

\paragraph{Reasoning-consistency penalties.}
We subtract $P_{\mathrm{motion}}=0.05$ when the first reasoning sentence omits, cannot express, or contradicts the mechanically recorded recent motion. A correct \texttt{STOP} whose reasoning still states that navigation is incomplete receives $P_{\mathrm{stop}}=0.50$. Finally, let $\rho_{\mathrm{cross}}$ be the fraction of unique current 4-grams also present in the previous decision's reasoning, after masking markup. If $\rho_{\mathrm{cross}}\ge0.85$, the final reward is overridden to $-1$.

\paragraph{Action-conditional composition.}
Let $H_t$ denote a gold-\texttt{MOVE}, same-family prediction whose target is either physically non-navigable or outside the front panel. For all other valid outputs, the base reward is
\begin{equation}
\widetilde R_t=
\begin{cases}
0.30R_{\mathrm{act}}+0.20R_{\mathrm{task}}+0.50R_{\mathrm{goal}},
& \bar a_t=\texttt{MOVE},\\
0.65R_{\mathrm{act}}+0.20R_{\mathrm{task}}+0.15\widetilde R_{\mathrm{goal}},
& \bar a_t=\texttt{TURN}\ \text{with target},\\
0.80R_{\mathrm{act}}+0.20R_{\mathrm{task}},
& \bar a_t=\texttt{TURN}\ \text{without target, or }\bar a_t=\texttt{STOP},
\end{cases}
\label{eq:gear-action-conditional-reward}
\end{equation}
where $\widetilde R_{\mathrm{goal}}=0$ for a same-family \texttt{TURN} with the wrong direction; otherwise it equals Equation~\ref{eq:gear-goal-reward}, including $-1$ for an ineligible target. The complete scalar reward is
\begin{equation}
R_t=
\begin{cases}
-1, & \text{invalid output or }\rho_{\mathrm{cross}}\ge0.85,\\
\operatorname{clip}_{[-1,1.1]}(-0.50-P_{\mathrm{grd}}), & H_t,\\
\operatorname{clip}_{[-1,1.1]}(\widetilde R_t+B_{\mathrm{grd}}
-P_{\mathrm{motion}}-P_{\mathrm{stop}}-P_{\mathrm{grd}}), & \text{otherwise}.
\end{cases}
\label{eq:gear-complete-reward}
\end{equation}
This ordering preserves the absolute $-0.50$ penalty for a physically invalid or side-view \texttt{MOVE}, while allowing the grounding penalty and the final cross-step-copy override to remain effective.

\subsection{Group-Relative Optimization Details}
\label{sec:appendix-grpo}

Following GRPO~\citep{shao2024deepseekmath}, for each anchor $x$, we sample $G$ outputs $\{y_i\}_{i=1}^{G}$ with rewards $\{r_i\}_{i=1}^{G}$. Their group-relative advantages are
\begin{equation}
    A_i=w_x\frac{r_i-\bar r_x}{\max(\sigma_x,\epsilon_A)},
    \label{eq:group-advantage}
\end{equation}
where $\bar r_x$ and $\sigma_x$ are the group mean and standard deviation. We set $w_x=0$ for groups in which every output is invalid or $\sigma_x$ falls below a variance threshold; groups containing valid reward variation but no correct decision receive a reduced weight $w_{\mathrm{wrong}}$, and all remaining groups use $w_x=1$.

Let $y_{i,j}$ be the $j$-th token of output $y_i$ and define the token-level importance ratio
$\rho_{i,j}(\theta)=\pi_\theta(y_{i,j}\mid x,y_{i,<j})/\pi_{\theta_{\mathrm{old}}}(y_{i,j}\mid x,y_{i,<j})$.
To regularize the policy without evaluating the full vocabulary distribution, we use a sampled-token KL proxy. Defining
$\Delta_{i,j}=\log\pi_{\mathrm{SFT}}(y_{i,j}\mid x,y_{i,<j})-\log\pi_\theta(y_{i,j}\mid x,y_{i,<j})$, the proxy is
\begin{equation}
    \widehat D_{i,j}^{\mathrm{KL}}
    =\exp(\Delta_{i,j})-\Delta_{i,j}-1.
    \label{eq:sampled-token-kl}
\end{equation}
The complete GEAR optimization objective is
\begin{equation}
\begin{aligned}
\ell_{i,j}^{\mathrm{clip}}(\theta)
&=\min\Big(
\rho_{i,j}(\theta)A_i,\\[-0.2em]
&\qquad\operatorname{clip}(\rho_{i,j}(\theta),1-\epsilon,1+\epsilon)A_i
\Big),\\
\mathcal{L}_{\mathrm{GEAR}}(\theta)
&=-\mathbb{E}_{x,\{y_i\}\sim\pi_{\theta_{\mathrm{old}}}}
\left[\frac{1}{G}\sum_{i=1}^{G}\frac{1}{|y_i|}
\sum_{j=1}^{|y_i|}\ell_{i,j}^{\mathrm{clip}}(\theta)\right]\\
&\quad+\beta\,\mathbb{E}_{x,\{y_i\},j}
\left[\widehat D_{i,j}^{\mathrm{KL}}\right].
\end{aligned}
\label{eq:gear-objective}
\end{equation}
where $\epsilon$ is the policy-ratio clipping range, $\beta$ controls sampled-token KL regularization, and $\pi_{\mathrm{SFT}}$ is the frozen SFT reference policy. Table~\ref{tab:gear-optimization-params} lists the optimization parameters used in all experiments.

\begin{table}[t]
\centering
\caption{GEAR optimization hyperparameters.}
\label{tab:gear-optimization-params}
\begingroup
\small
\setlength{\tabcolsep}{4pt}
\renewcommand{\arraystretch}{1.02}
\begin{tabular}{@{}p{0.65\columnwidth}c@{}}
\toprule
Parameter & Value \\
\midrule
Sampled outputs per anchor $G$ & $8$ \\
Advantage standard-deviation floor $\epsilon_A$ & $0.05$ \\
Low-variance group threshold $\tau_{\mathrm{var}}$ & $0.02$ \\
All-wrong group weight $w_{\mathrm{wrong}}$ & $0.40$ \\
Policy-ratio clipping range $\epsilon$ & $0.20$ \\
Sampled-token KL coefficient $\beta$ & $0.02$ \\
Learning rate & $1\!\times\!10^{-6}$ \\
Training steps & $2{,}000$ \\
\texttt{MOVE}/\texttt{TURN}/\texttt{STOP} sampling ratio & $0.50/0.30/0.20$ \\
\bottomrule
\end{tabular}
\endgroup
\end{table}

\subsection{Limitations}
\label{sec:appendix-limitations}

Although the high-level VLM in GroundingVLN operates solely on RGB observations, the low-level execution module currently relies on sensor-provided metric depth to translate pixel goals into precise and reliable actions. Without training an additional vision-only point-goal planner, depth is required for 2D-to-3D back-projection and RGB-D SLAM-based pose and map estimation. This dependency limits deployment on platforms without reliable depth sensing, motivating future work on an RGB-only low-level planner that does not rely on metric depth while preserving geometric accuracy.

\end{document}